\documentclass[12pt]{article}

\usepackage{sbc-template}

\usepackage{graphicx,url}
\usepackage{float}
\usepackage[brazil]{babel}   
\usepackage[utf8]{inputenc}  
\usepackage{tabularx} 
\usepackage{ragged2e} 
\usepackage{booktabs} 
\usepackage[table]{xcolor} 
\usepackage{graphicx} 
\usepackage{placeins}

\definecolor{gold}{RGB}{106,81,12}
\definecolor{silver}{RGB}{90,85,85}
\definecolor{bronze}{RGB}{171,14,14}

\newif\ifanonymous
\anonymousfalse  

\newcommand{\anonimizar}[2]{%
  \ifanonymous
    {\color{black}\textit{[Dados Omitidos para Revisão]}}%
  \else
    #2%
  \fi
}

\newcommand{\anonimizarfootnote}[2]{%
  \ifanonymous
    {\color{black}\textit{[Dados Omitidos para Revisão]}\footnote{\color{black}\textit{[Nota omitida para revisão]}}}%
  \else
    #1\footnote{#2}%
  \fi
}

\newcolumntype{L}{>{\RaggedRight\arraybackslash}X} 
\newcolumntype{C}{>{\Centering\arraybackslash}X} 
\newcolumntype{R}{>{\RaggedLeft\arraybackslash}X} 

\title{Sabiá: Um Chatbot de Inteligência Artificial Generativa para Suporte no Dia a Dia do Ensino Superior}

\author{%
  \anonimizar{}{Guilherme Biava Rodrigues\inst{1}, Franciele Beal\inst{2}, Marlon Marcon\inst{1} \\
  Alinne Cristinne Corrêa Souza\inst{1}, André Roberto Ortoncelli\inst{1}\\Francisco Carlos Monteiro Souza\inst{1} Rodolfo Adamshuk Silva\inst{1}}%
}

\address{%
  \anonimizar{}{%
    Coordenação de Engenharia de Software - COENS-DV\\
    Universidade Tecnológica Federal do Paraná - Campus Dois Vizinhos (UTFPR)\\
    Estr. p/ Boa Esperança, s/n - KM 04 - Zona Rural, Dois Vizinhos - PR - Brasil
    \nextinstitute
    Departamento Acadêmico de Informática - DAINF-PB\\
    Universidade Tecnológica Federal do Paraná - Campus Pato Branco (UTFPR)\\
    Via do Conhecimento, s/n - KM 01 - Fraron, Pato Branco - PR - Brasil
    \email{guilhermebiavarodrigues@alunos.utfpr.edu.br}
    \email{\{fbeal, marlonmarcon, alinnesouza, ortoncelli\}@utfpr.edu.br}
    \email{\{franciscosouza, rodolfoa\}@utfpr.edu.br}
  }%
}

\begin{document} 

\maketitle

\begin{abstract}
  Students often report difficulties in accessing day-to-day academic information, which is usually spread across numerous institutional documents and websites. This fragmentation results in a lack of clarity and causes confusion about routine university information. This project proposes the development of a chatbot using Generative Artificial Intelligence (GenAI) and Retrieval-Augmented Generation (RAG) to simplify access to such information. Several GenAI models were tested and evaluated based on quality metrics and the LLM-as-a-Judge approach. Among them, Gemini 2.0 Flash stood out for its quality and speed, and Gemma 3n for its good performance and open-source nature.
\end{abstract}
     
\begin{resumo} 
Estudantes frequentemente relatam dificuldades no acesso a informações do cotidiano acadêmico, geralmente dispersas em muitos documentos institucionais e websites. Essa fragmentação leva a falta de clareza e confusão sobre informações do dia a dia universitário. Este projeto propõe o desenvolvimento de um chatbot, utilizando Inteligência Artificial Generativa (GenAI) e Geração Aumentada por Recuperação (RAG), para simplificar o acesso a essas informações. Diversos modelos de GenAI foram testados e avaliados com base em métricas de qualidade e com a abordagem LLM-as-a-Judge. Entre eles, destacaram-se o Gemini 2.0 Flash, por sua qualidade e velocidade, e o Gemma 3n, que além de um bom desempenho, tem natureza opensource.

\end{resumo}

\section{Introdução}
No dia a dia das instituições de ensino, é comum que docentes e discentes necessitem consultar informações relacionadas ao exercício de suas atividades acadêmicas. Tais informações abrangem desde documentos voltados ao trabalho docente, como normativas internas, regulamentos institucionais, resoluções, ofícios e editais, até aspectos da vida acadêmica discente, como procedimentos para matrícula, regras de estágio supervisionado, orientações sobre Trabalho de Conclusão de Curso (TCC), aproveitamento de disciplinas, prazos e formulários diversos.

As formas mais comuns de acesso a essas informações incluem a navegação nos portais institucionais, buscas na web, o envio de e-mails para departamentos administrativos ou coordenações de curso, além da comunicação direta com professores e colegas. No entanto, tais estratégias apresentam diversas limitações. Frequentemente, os portais institucionais são desorganizados, apresentam documentos desatualizados, \textit{links} quebrados e carecem de mecanismos de busca eficientes. Além disso, as informações costumam estar dispersas em diferentes páginas ou documentos extensos e técnicos, exigindo esforço adicional por parte do usuário para localizar e compreender os procedimentos acadêmicos.

A Inteligência Artificial Generativa (GenAI) tem se consolidado como uma tecnologia amplamente utilizada no cotidiano das pessoas, especialmente como ferramenta para tirar dúvidas, aprender novos conhecimentos e facilitar tarefas diversas. Pesquisa como de \cite{naik2024} descrevem que a GenAI aprimora o acesso dos estudantes às informações institucionais ao potencializar os \textit{chatbots} que fornecem respostas precisas e oportunas às dúvidas acadêmicas. Essas interfaces inteligentes suportam entradas de texto e voz, garantindo interações fáceis de usar e recursos multilíngues para diversas necessidades dos alunos \cite{naik2024}. Para \cite{noviandy2024}, \textit{chatbots} com tecnologia de GenAI podem lidar com eficiência com as dúvidas dos estudantes, auxiliar em tarefas administrativas de rotina, tornando as matrículas em cursos e a orientação acadêmica mais simplificadas e acessíveis \cite{noviandy2024}. 

Diante desse cenário, este trabalho buscou responder à seguinte questão de pesquisa: \textit{a GenAI pode auxiliar estudantes no acesso a informações institucionais e na resolução de dúvidas relacionadas à vida acadêmica, atuando como uma interface de consulta inteligente? Ao ser treinada com documentos institucionais, como regulamentos, editais, calendários e instruções normativas, seria possível que ela oferecesse respostas rápidas, objetivas e assertivas aos estudantes?}

Para investigar essa possibilidade, foi desenvolvida uma GenAI denominada Sabiá, projetada para apoiar a comunidade universitária em tarefas acadêmicas e administrativas, otimizando o acesso à informação e promovendo a eficiência de processos internos. A proposta consiste em criar um \textit{chatbot} especializado, capaz de compreender e interpretar documentos institucionais da \anonimizar{}{Universidade Tecnológica Federal do Paraná (UTFPR)}, fornecendo respostas em linguagem natural e acessível.

O nome adotado para o assistente virtual desenvolvido, Sabiá, se deve ao fato dessa ave ser conhecida como a Ave Nacional do Brasil \cite{decreto9675_2002}. O Sabiá é amplamente conhecido entre os brasileiros pelo seu canto melodioso, que é frequentemente associado à comunicação. Inspirado nessas características, o nome "Sabiá" foi escolhido como uma forma de valorizar esse símbolo cultural brasileiro.

A justificativa deste trabalho reside em um problema recorrente no ambiente universitário: o excesso de burocracia, a linguagem excessivamente técnica dos documentos e a fragmentação da informação. Tais barreiras impactam diretamente a experiência de docentes e discentes, gerando dúvidas frequentes, perda de prazos, retrabalho e desinformação. Nesse contexto, o Sabiá surge como uma solução prática e inclusiva, oferecendo atendimento inteligente e disponível 24 horas por dia. Desta forma, acredita-se que as principais contribuições deste trabalho são:
\begin{itemize}
    \item O desenvolvimento de uma solução tecnológica que promove o acesso facilitado e equitativo à informação acadêmica e administrativa no ambiente universitário. Ao oferecer uma IA treinada com dados institucionais, o projeto contribui para a redução de barreiras informacionais, beneficiando estudantes e professores. 
    \item A disponibilização da solução com detalhes de sua implementação, permitindo que outras instituições possam adaptá-la e integrá-la aos seus contextos. Dessa forma, o projeto colabora com o processo de transformação digital das universidades públicas brasileiras, aproximando a experiência institucional dos padrões tecnológicos contemporâneos.
\end{itemize}

\section{Chatbots, IA Generativa e a Geração Aumentada por Recuperação (RAG)}

Do ponto de vista conceitual, \textit{chatbots}, também conhecidos como assistentes virtuais ou apenas \textit{bots}, são projetos de software que tem como intuito responder a questionamentos de usuários e simular diálogos em linguagem natural, a fim de tornar a interação humano-computador mais realista e fluída. Atualmente, onde o acesso à informação é requisitado em um espaço de tempo cada vez menor, soluções informatizadas como os \textit{chatbots} podem automatizar e diminuir a necessidade de comunicação direta com um humano, garantindo uma maior velocidade no acesso às informações \cite{Carvalho2018}.

Em sua forma mais tradicional, os \textit{chatbots} costumam operar em fluxos conversacionais estruturados de uma forma rígida, na qual o usuário escolhe entre opções predefinidas e o \textit{bot} responde com base nesses caminhos previamente programados. Apesar de funcional, essa abordagem apresenta limitações quando se busca uma experiência mais imersiva e natural na interação com estes assistentes \cite{Adamopoulou2020}. Técnicas como o Processamento de Linguagem Natural (PLN), a IA Generativa (GenAI) e a Geração Aumentada por Recuperação (RAG) podem auxiliar na geração de respostas mais naturais e melhorar significativamente a experiência de usuários. Um exemplo representativo dessa nova geração de \textit{chatbots} é o ChatGPT\footnote{ChatGPT - https://chatgpt.com}, que permite ao usuário manter uma conversa fluida com um agente de IA por meio de \textit{prompts} em linguagem natural, sem interações previamente roteirizadas.

A Inteligência Artificial Generativa (GenAI), por sua vez, consiste em uma tecnologia capaz de gerar conteúdos originais, como imagens, vídeos, textos e trechos de códigos, com base em um conhecimento adquirido por meio de um treinamento com grandes conjuntos de dados. \cite{Carvalho2024}. 

Embora os conteúdos produzidos por sistemas de GenAI à primeira vista pareçam bem escritos e coerentes, essa impressão pode ser enganosa em alguns casos. Esses sistemas são capazes de gerar textos que soam naturais e convincentes, mas isso não significa que eles realmente compreendam o que estão dizendo ou que as informações geradas sejam adaptadas a todos os contextos. Como alertam \cite{Bender2021} em seu estudo, modelos de GenAI operam com base em padrões estatísticos aprendidos durante o treinamento, e não em entendimento real do conteúdo. Por isso, podem produzir respostas incorretas ou inventadas, comumente conhecidas como "alucinações".

Tendo isso em vista, a Geração Aumentada por Recuperação (RAG) surge como uma abordagem híbrida na qual modelos de GenAI são integrados a módulos de recuperação de informações, permitindo que as respostas geradas sejam baseadas tanto no conhecimento pré-treinado, quanto em informações externas recuperadas dinamicamente \cite{Lewis2020}. 

A estrutura básica de um sistema RAG envolve dois principais componentes: um mecanismo de recuperação (\textit{retriever}), que localiza trechos de texto relevantes a partir de uma consulta; e um modelo gerador (\textit{generator}) de LLM, que utiliza essas informações recuperadas para produzir uma resposta coerente e enriquecida com base nas fontes buscadas \cite{Gupta2024}. Essa combinação de técnicas, permite que o modelo supere limitações de desatualização e reduza alucinações que são comuns em modelos de GenAI puramente generativos, isso torna o RAG uma abordagem ideal para aplicações que exigem uma maior precisão e contextualização de informações.

Nos últimos tempos, essa arquitetura vem ganhando cada vez mais espaço em contextos práticos, justamente por se diferenciar de modelos de IA puramente generativos, que dependem apenas do conhecimento adquirido durante processos de treinamento. Como o RAG busca dados externos em fontes específicas — como bases de conhecimento, documentos, bancos de dados ou APIs — no momento da geração das respostas, ele é capaz de fornecer uma resposta mais confiável e personalizada, evitando alucinações, ao mesmo tempo em que permite a adaptação do sistema às necessidades específicas de uma aplicação prática \cite{Siriwardhana2023}.

\section{Trabalhos relacionados}
Recentemente, diversos estudos têm explorado como o uso de assistentes virtuais e técnicas de IA podem apoiar o dia a dia acadêmico, buscando melhorar o acesso a informações e melhorar a experiência dos estudantes. A seguir, alguns trabalhos são destacados por servirem de referência ao desenvolvimento do Sabiá.

No trabalho de \cite{mendes2020}, um \textit{chatbot} é desenvolvido para o acolhimento de novos estudantes nas instituições de ensino superior. O foco consiste em orientações iniciais para os novos discentes, como localização de salas de aula e horários. A aplicação, baseada em técnicas de PLN, evidencia como essas abordagens podem proporcionar maior eficiência em interações recorrentes da vida acadêmica. Diferentemente dessa proposta, o Sabiá amplia o escopo ao não se restringir a fluxos fixos de acolhimento, oferecendo interações mais naturais e adaptáveis por meio de GenAI, além de abranger diferentes demandas acadêmicas que vão além da recepção de novos estudantes.

Em um contexto semelhante, o estudo apresentado por \cite{Neto2020} propõe um \textit{chatbot} para responder dúvidas dos usuários relacionadas ao Catálogo de Cursos da Rede Federal de Educação Profissional, Científica e Tecnológica. Assim como no trabalho de \cite{mendes2020}, a solução utiliza PLN para permitir consultas em linguagem natural e reforça a importância de uma comunicação flexível. Embora compartilhe esse objetivo, o Sabiá se diferencia ao explorar GenAI, superando a limitação de um domínio específico e possibilitando maior adaptabilidade e replicabilidade em diferentes contextos institucionais.

Por outro lado, trabalhos mais recentes, como o de \cite{Carvalho2024}, utilizam diretamente modelos de GenAI. Nesse estudo, os autores propõem uma aplicação com a geração automática de \textit{quizzes} personalizados por meio de uma API do modelo \textit{Gemini}\footnote{Gemini - https://deepmind.google/models/gemini/}, auxiliando os estudantes na fixação de conteúdos. A facilidade de implementação e de replicabilidade da aplicação serviram como uma das principais inspirações para este trabalho, além de demonstrar o potencial de soluções de GenAI no contexto da educação.

Além das contribuições dos trabalhos analisados, este projeto também se inspira em soluções já utilizadas por instituições de ensino, como o MinhaUFC\footnote{MinhaUFC - https://www.ufc.br/a-universidade}, assistente virtual da Universidade Federal do Ceará. Embora funcione como um \textit{chatbot} tradicional, com interações pré-programadas, esta ferramenta se destaca por fornecer informações de diversos contextos do dia a dia dos acadêmicos, além de estar facilmente acessível no site da universidade, por meio de um ícone discreto no canto da tela. Essa forma de integração permite que os estudantes encontrem ajuda com facilidade, sempre que necessário. No entanto, a limitação nas possibilidades de interação — restritas a perguntas previstas — compromete a naturalidade da conversa e. Diante disso, o Sabiá se propõe a superar essas restrições, unindo a acessibilidade e disponibilidade das informações com o poder da GenAI, lidando com fluxos naturais de interação em vez de respostas previamente programadas, além de adaptar e personalizar suas respostas de acordo com diferentes contextos, mantendo a fluidez e a fácil implementação por meio de APIs.

\section{Metodologia}

Para desenvolver o Sabiá, o trabalho foi estruturado em quatro etapas, que serão exploradas na sequência, sendo elas: o levantamento de requisitos do sistema, com base nas perguntas que motivaram o projeto (etapa 1); o desenvolvimento da aplicação com foco na facilidade de implementação, replicação e customização do trabalho (etapa 2); a coleta e preparação dos documentos institucionais utilizados como base para a recuperação de informações (etapa 3); e por último, a definição e aplicação de critérios para avaliação da qualidade das respostas geradas (etapa 4).

\subsection{Levantamento de Requisitos - Etapa 1}
Os requisitos para o desenvolvimento do Sabiá foram definidos tendo como base as questões que motivaram este estudo: \textit{seria possível utilizar a inteligência artificial generativa para melhorar o acesso a informações institucionais e auxiliar estudantes na resolução de dúvidas sobre a vida acadêmica?} Como também, \textit{A IA, ao ser treinada com documentos como regulamentos seria capaz de oferecer respostas rápidas, claras e relevantes?} Com base nessas perguntas, foram definidos alguns requisitos funcionais que guiassem à criação de um assistente virtual eficaz. Os requisitos levantados podem ser observados na Tabela \ref{tab:requisitos_funcionais}.

\begin{table}[htbp]
    \centering
    \caption{Requisitos Funcionais}
    \label{tab:requisitos_funcionais}
    \footnotesize
    \begin{tabularx}{\columnwidth}{l X}
        \toprule
        \textbf{Código} & \textbf{Descrição} \\
        \midrule

        RF001 & Permitir fácil integração com diferentes modelos de linguagem natural. \\
        \addlinespace

        RF002 & Recuperar informações com base em documentos oficiais da universidade. \\
        \addlinespace

        RF003 &  Responder a dúvidas relacionadas ao cotidiano acadêmico, como estágio, TCC entre outros. \\
        \addlinespace

        RF004 & Apresentar respostas com clareza e objetividade. \\
        \addlinespace

        RF005 & Ser acessível por meio de navegadores web. \\
        
        \bottomrule
    \end{tabularx}
\end{table}

\subsection{Desenvolvimento do Sabiá - Etapa 2}
\begin{figure}[ht]
\centering
\includegraphics[width=.70\textwidth, trim=0cm 1.5cm 0cm 1.5cm, clip]{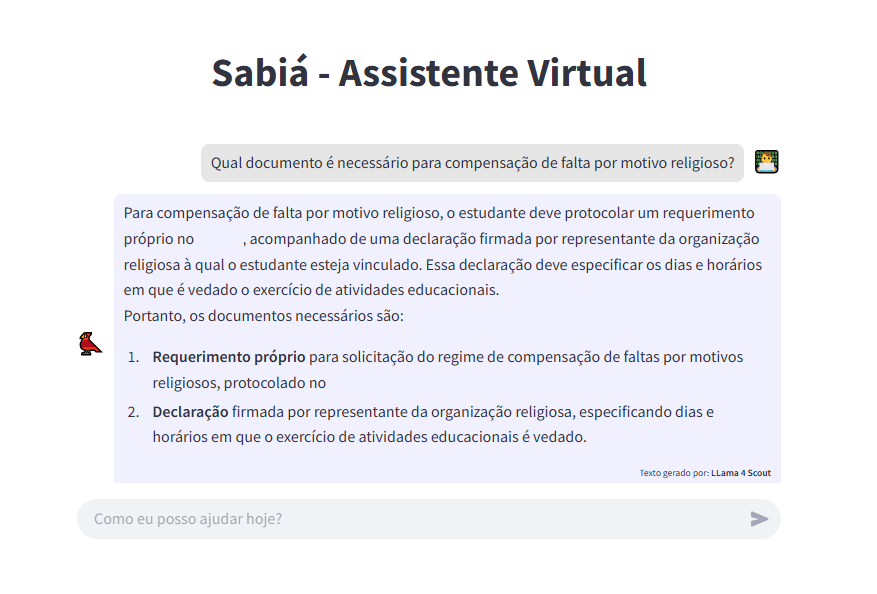}
\caption{Exemplo da aplicação.}
\label{fig:Aplicacao}
\end{figure}

O Sabiá é uma aplicação web, proposta para ter uma facilidade na sua replicação e personalização. O objetivo da proposta é fornecer uma interface onde os estudantes podem interagir de maneira fluída e natural para sanarem suas dúvidas e questões do dia a dia universitário. A implementação se diferencia de \textit{chatbots} convencionais com estruturas pré-definidas, pois combina o uso de \textit{GenAI} para geração de respostas naturais e do RAG para permitir que o sistema busque informações diretamente em documentos institucionais e contextualize as saídas. A Figura \ref{fig:Aplicacao} mostra um exemplo de aplicação do Sabiá, onde um estudante faz a pergunta \textit{Qual documento é necessário para compensação de falta por motivo religioso?} e obtém a resposta gerada pelo LLM LLama 4 Scout.

\begin{figure}[ht]
\centering
\includegraphics[width=.60\textwidth]{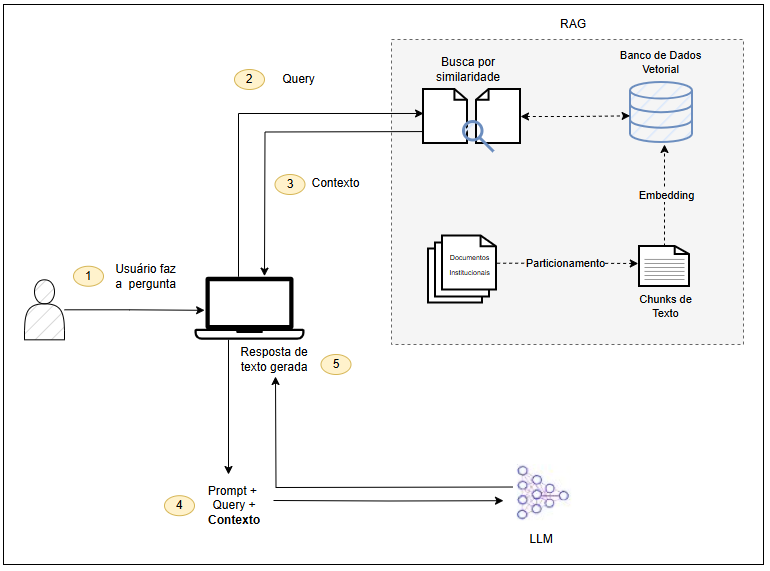}
\caption{Estrutura da Aplicação.}
\label{fig:estrutura}
\end{figure}

A arquitetura do sistema, apresentada na Figura \ref{fig:estrutura}, é composta por partes funcionais bem definidas. A figura ilustra o fluxo completo da aplicação, que será descrito ao longo desta seção. A interface da aplicação (\textit{frontend}) foi desenvolvida em Python utilizando o framework \textit{Streamlit}\footnote{Streamlit - https://streamlit.io/}, o que permitiu uma criação ágil da interface visual interativa. A escolha desta tecnologia se deu graças à sua curva de aprendizado baixa e à integração eficiente com bibliotecas voltadas a inteligência artificial. O Sabiá pode ser acessado por um navegador web e a sua página principal consiste em um campo de entrada para as perguntas do usuário, um histórico de mensagens trocadas entre o assistente virtual e o estudante e um menu lateral para seleção dos LLMs disponíveis. 

No \textit{backend}, a aplicação baseia-se em um \textit{pipeline} RAG com o uso da biblioteca \textit{Langchain}\footnote{Langchain - https://www.langchain.com/}, que permite a orquestração entre a recuperação de informação dos documentos institucionais armazenados na aplicação como também a geração de respostas por meio de modelos de linguagem. 

Antes de iniciar o processo do \textit{pipeline} RAG, é realizada uma etapa de preparação dos dados, responsável por realizar o pré-processamento dos documentos em PDF, extraindo seus conteúdos e dividindo-os em trechos (\textit{chunks}). Após este particionamento, os trechos são convertidos em \textit{embeddings}, que consistem em representações vetoriais de segmentos de texto, capazes de armazenar o significado semântico de cada fragmento textual. Posteriormente, esses \textit{embeddings} são armazenados em um banco de dados vetorial, utilizando o modelo \textit{opensource} \textit{ChromaDB}\footnote{ChromaDB - https://docs.trychroma.com/docs/overview/introduction}. O uso de um banco de dados vetorial é crucial, pois ele permite que seja realizada a busca por similaridade semântica — ou seja, buscar trechos cujo conteúdo seja mais próximo ao significado da pergunta feita pelo usuário, mesmo que as palavras não sejam exatamente as mesmas.


Quando o usuário envia uma pergunta pela interface (passo 1), o sistema inicia a recuperação de contexto. A aplicação envia o \textit{prompt} do usuário, que é convertido em vetor pelo pipeline RAG (passo 2) e comparado com os vetores de documentos institucionais armazenados no banco de dados vetorial. Com base nessa busca por similaridade, os resultados mais relevantes são ranqueados e os trechos com maior pontuação são selecionados para formar um contexto (passo 3), que é adicionado à pergunta original para tornar a resposta do LLM mais relevante, precisa e alinhada ao conteúdo institucional.


O contexto recuperado e a pergunta original são enviados ao LLM junto de um \textit{template} que orienta o tom, idioma e comportamento frente a dúvidas não respondidas (passo 4). Isso caracteriza a abordagem RAG, que introduz a recuperação de informações externas antes da geração da resposta, permitindo ao LLM combinar seu treinamento com conteúdos atualizados dos documentos institucionais, tornando as respostas mais contextuais e precisas.

Para a geração das respostas, foram integrados diversos modelos de LLMs por meio da plataforma \textit{OpenRouter}\footnote{Open Router - https://openrouter.ai/}, que permite a geração de \textit{tokens} de \textit{API} de forma gratuita para o uso desses modelos. O LLM escolhido recebe o contexto, a pergunta original e as instruções do \textit{template}, e, com base nisso, gera a resposta textual (passo 5). Dentre os LLMs selecionados, destacam-se os modelos de código aberto (do inglês, \textit{OpenSource}) \textit{DeepSeek R1}, \textit{LLaMA 4 Scout}, \textit{Gemma-3n}, \textit{Phi-4-Reasoning} e \textit{Qwen3-235bo}. Além destes, modelos prioritários como \textit{GPT-4o-mini} e \textit{Gemini 2.0 Flash} também foram adotados.

A implementação de diversos modelos de LLMs permite avaliar o desempenho das respostas geradas por diferentes arquiteturas de IA Generativa, analisando seu desempenho com base em critérios definidos na Seção \ref{sec:avaliacao}. O objetivo é identificar qual modelo oferece as respostas mais adequadas ao contexto acadêmico, considerando aspectos como relevância, acurácia entre outros, com especial atenção ao potencial dos modelos de código aberto, que se destacam pela possibilidade de uso gratuito.


É importante destacar que o repositório do projeto completo está disponibilizado publicamente por meio da URL: {\textbf{\url{https://github.com/guilhermebiava/Sabia}}, permitindo que qualquer pessoa possa clonar para utilizar, adaptar ou expandir o sistema conforme suas necessidades. A única exigência para seu funcionamento é a criação de uma chave de API para acesso aos modelos desejados, seja via \textit{OpenRouter}, como utilizado ao longo do projeto, ou através de outras plataformas que gerem os \textit{tokens} de API, executando em servidores dedicados ou mesmo localmente.

\subsection{Levantamento da Base de Dados - Etapa 3}

A terceira etapa no desenvolvimento da aplicação consistiu na identificação e seleção dos documentos institucionais mais relevantes para compor a base de conhecimento a ser recuperada pelo RAG. Foram priorizados arquivos em formato PDF que contêm informações frequentemente consultadas por estudantes, como \anonimizar{}{Regulamento de Organização Didático-Pedagógica (RODP)}, regulamentos de Trabalho de Conclusão de Curso (TCC), de estágio, regulamentos acerca de Atividades Complementares e documentos complementares disponíveis no site oficial da \anonimizarfootnote{Universidade}{UTFPR - Diretrizes e regulamentos - https://www.utfpr.edu.br/documentos/} e no site oficial da \anonimizarfootnote{Coordenação do Curso de Engenharia de \textit{Software}}{Engenharia de Software - https://coens.dv.utfpr.edu.br/site/}. Por se tratarem de documentos institucionais e de acesso público, a manipulação desses materiais não envolve dados pessoais, o que reduz preocupações relacionadas à conformidade com a Lei Geral de Proteção de Dados (LGPD). Além disso, apenas arquivos compostos por texto e tabelas foram considerados, sendo descartados documentos cujo conteúdo era predominantemente imagético. Esses materiais foram então manipulados e empregados na etapa de criação do banco vetorial da aplicação, como citado anteriormente.

\subsection{Avaliação - Etapa 4}
\label{sec:avaliacao}

Para testar e avaliar o Sabiá, foi utilizada uma lista de perguntas que foram elaboradas e respondidas pela \anonimizar{}{coordenação do curso de Bacharelado em Engenharia de Software da UTFPR do campus Dois Vizinhos} e pelo(a) professor(a) responsável pelos estágios, com base em dúvidas reais e recorrentes dos estudantes. Neste trabalho, essa lista foi chamada de FAQ (do inglês, \textit{Frequently Asked Questions}). Além de conter as perguntas e respostas do FAQ, um arquivo detalhado que consta também com as respostas geradas pelos LLMs pode ser acessado por meio do seguinte link: {\textbf{\url{https://github.com/guilhermebiava/Sabia}}. Este foi disponibilizado na pasta \textit{apps}, e consiste no arquivo CSV \textit{resultadosmodelos.csv}. As perguntas da FAQ foram utilizadas como \textit{prompts} (entradas) para os modelos, enquanto as respectivas respostas serviram de referência para avaliação quanto a adequação e qualidade das saídas geradas.

\subsubsection{Métricas de Avaliação}


A avaliação de textos gerados por LLMs requer métricas que analisem não apenas a aparência dos textos (como estrutura e vocabulário), mas também o significado das informações comunicadas. Este estudo utiliza abordagens tradicionais e emergentes, destacando suas aplicações e limitações no contexto educacional. A seguir são apresentadas tais métricas:

\begin{itemize}

    \item \textbf{ROUGE} (\textit{Recall-Oriented Understudy for Gisting Evaluation}): usada para avaliar resumos automáticos comparando trechos do texto gerado com trechos de um texto de referência e medindo quantas partes se repetem. Variantes como o ROUGE-N (como ROUGE-1, ROUGE-2) mede a sobreposição de  \textit{n-gramas} (sequências contíguas de \textit{n} palavras), e o ROUGE-L que identifica a sequência mais longa de palavras comum entre os textos. Essa métrica verifica se o conteúdo principal foi mantido. Sua pontuação varia de 0 a 1, onde valores mais altos indicam uma melhor qualidade do resumo.\cite{rao2025evaluation}.


     \item \textbf{BLEU} (\textit{Bilingual Evaluation Understudy}): originalmente usada para avaliar a qualidade de traduções automáticas, essa métrica mede a similaridade entre o texto gerado e o texto de referência usando \textit{n-gramas}. Porém, ela tem dificuldade em lidar com textos mais criativos, que podem usar palavras diferentes para expressar as mesmas ideias. Apesar disso, é fácil de aplicar e funciona para diferentes idiomas. Sua pontuação varia de 0 a 1, onde valores mais altos indicam melhor qualidade da tradução. \cite{sarawanangkoor2024}.



    \item \textbf{SBERT} \textit{(Sentence-BERT)}, introduzido por \cite{reimers2019sentence}, é uma modificação da arquitetura BERT (Bidirectional Encoder Representations from Transformers) utilizada para comparar o significado de sentenças, verificando se elas transmitem a mesma ideia, mesmo com palavras diferentes. Para isso, transforma cada sentença em um vetor numérico chamado \textit{embedding}, que captura o sentido da sentença. Sentenças com significados parecidos geram vetores próximos entre si, permitindo comparar textos com base no sentido, e não apenas nas palavras usadas. A pontuação varia de -1 (significados opostos) a 1 (significados semelhantes).


    \item \textbf{METEOR} \textit{(Metric for Evaluation of Translation with Explicit ORdering)} foi criada para superar limitações do BLEU, calculando o alinhamento entre palavras do texto gerado e do texto de referência. Esse alinhamento considera correspondências exatas, de radicais (via \textit{stemming}), sinônimos (usando recursos como o WordNet para o inglês) e, em algumas versões, também paráfrases \cite{banerjee2005meteor}.
    
    \item \textbf{\textit{LLM-as-a-Judge}}: emprega LLMs como avaliadores automáticos. Estudos mostram que essa técnica tem um alto grau de concordância com avaliações humanas (com correlação acima de 0,85). Técnicas como \textit{Reference-Guided Verdict} automatizam avaliações de tarefas abertas, capturando nuances contextuais perdidas por métricas tradicionais. A pontuação retornada ao final de uma tarefa é um índice de 0 a 1, que corresponde ao índice de similaridade.  \cite{badshah2024reference}.
    
\end{itemize}

\vspace{-1cm}
\subsubsection{Processo de Avaliação}

O processo de avaliação inciou com a execução dos \textit{prompts} nos modelos candidatos. As respostas dos LLMs (\textit{saída}) foram coletadas e submetidas às métricas de avaliação citadas anteriormente, para serem comparadas com as respostas de referência da lista FAQ. 

O LLM utilizado como \textit{LLM-as-a-Judge} foi o "GPT-4.1-mini", diferente dos demais utilizados para geração das resposta. Para avaliar, o LLM recebeu as respostas da FAQ, as respostas dos modelos candidatos (\textit{Respostas dos LLMs}) e a rubrica de avaliação descrita na Tabela \ref{tab:rubrica_llm_avaliacao} que detalha os critérios para avaliar a qualidade das respostas geradas pelos LLM. Cada critério possui uma escala de 1 a 5, onde 1 indica desempenho insatisfatório e 5 indica desempenho excepcional. A Figura \ref{fig:avaliacao} ilustra o processo de avaliação.

\begin{figure}[h]
\centering
\includegraphics[width=.55\textwidth]{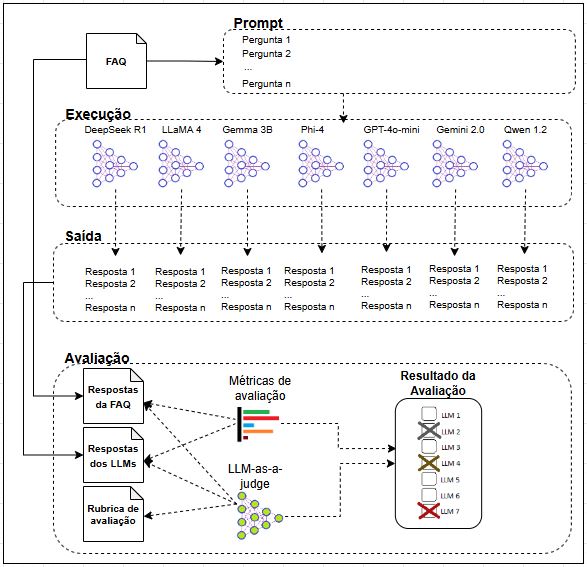}
\caption{Processo de Avaliação - Etapa 4.}
\label{fig:avaliacao}
\end{figure}


\begin{table*}[]
    \centering
    \caption{Rubrica de Avaliação para Respostas de LLMs}
    \label{tab:rubrica_llm_avaliacao}
    \footnotesize
    \setlength{\tabcolsep}{4pt} 
    \resizebox{0.9\textwidth}{!}{%
    \begin{tabularx}{\textwidth}{@{} l *{5}{X} @{}}
        \toprule
        \textbf{Critério} & \textbf{Insatisfatório} & \textbf{Parcialmente Satisfatório} & \textbf{Satisfatório} & \textbf{Bom} & \textbf{Excepcional} \\
        \midrule

        \textbf{Relevância} & Não aborda o tópico solicitado. & Aborda parcialmente com desvios significativos. & Aborda o tópico com alguns desvios menores. & Resposta direta e pertinente ao tópico. & Altamente focada e muito pertinente. \\
        \addlinespace
        
        \textbf{Acurácia} & Contém erros graves ou alucinações evidentes. & Apresenta imprecisões claras que prejudicam credibilidade. & Majoritariamente correta com pequenas imprecisões. & Informações corretas e verificáveis. & Muito acurada com informações confiáveis. \\
        \addlinespace
        
        \textbf{Completude} & Falha em cobrir aspectos essenciais. & Cobre apenas alguns aspectos necessários. & Cobre a maioria dos aspectos necessários. & Cobre todos os aspectos de forma abrangente. & Cobertura muito completa e detalhada. \\
        \addlinespace
        
        \textbf{Clareza} & Difícil de entender, mal organizada. & Dificuldades de compreensão, fluxo inconsistente. & Compreensível e razoavelmente organizada. & Clara, bem organizada, fluxo lógico. & Muito clara e bem estruturada. \\
        \addlinespace
        
        \textbf{Concisão} & Excessivamente longa e redundante. & Prolixa com informações supérfluas. & Razoavelmente concisa, alguns excessos. & Concisa e direta ao ponto. & Muito concisa e eficiente. \\
        \bottomrule
    \end{tabularx}
    }
\end{table*}

\vspace{-1cm}

\vspace{-1cm}

\vspace{-0.5cm}

\section{Resultados}

Esta seção apresenta os resultados da avaliação de desempenho dos diferentes LLMs integrados ao Sabiá. A análise foi conduzida com base nas métricas de qualidade de resposta e no tempo de processamento, conforme detalhado Seção \ref{sec:avaliacao}.

A avaliação da qualidade das respostas, sumarizada na Tabela~\ref{tab:llm_metrics_color}, apresenta o comparativo dos resultados das métricas que quantifica a comparação das saídas dos modelos com as respostas de referência. Utilizando a abordagem \textit{LLM-as-a-Judge} \cite{badshah2024reference}, que emprega um LLM para avaliar a qualidade das respostas geradas, os modelos que mais se destacaram foram o Phi 4, com uma pontuação de $0.768 \pm 0.26$, seguido de perto pelo Qwen3-235b, com $0.766 \pm 0.28$, e o DeepSeek R1, com $0.752 \pm 0.27$. Essa métrica avalia critérios como relevância, acurácia, completude, clareza e concisão, conforme a rubrica de avaliação apresentada na Tabela~\ref{tab:rubrica_llm_avaliacao}.

\vspace{-0.5cm}

\begin{table}[H]
\centering
\caption{Comparação de desempenho entre LLMs. Os valores são média $\pm$ desvio padrão. Os três melhores resultados são coloridos para indicar o 1º (\textcolor{gold}{ouro}), 2º (\textcolor{silver}{prata}) e 3º (\textcolor{bronze}{bronze}) lugar. O melhor resultado também está em \textbf{negrito}.}
\label{tab:llm_metrics_color}
\resizebox{\textwidth}{!}{%
\begin{tabular}{lccccccc}
\toprule
\textbf{Modelo} & \textbf{ROUGE-1} & \textbf{ROUGE-2} & \textbf{ROUGE-L} & \textbf{BLEU} & \textbf{SBERT} & \textbf{METEOR} & \textbf{LLM-as-a-Judge} \\
\midrule
GPT 4o & 0.387 $\pm$ 0.25 & 0.247 $\pm$ 0.27 & 0.362 $\pm$ 0.26 & 0.193 $\pm$ 0.25 & \textcolor{bronze}{0.734 $\pm$ 0.15} & 0.367 $\pm$ 0.27 & 0.741 $\pm$ 0.29 \\
DeepSeek R1 & 0.362 $\pm$ 0.26 & 0.223 $\pm$ 0.28 & 0.341 $\pm$ 0.26 & 0.164 $\pm$ 0.24 & 0.724 $\pm$ 0.15 & 0.340 $\pm$ 0.26 & \textcolor{bronze}{0.752 $\pm$ 0.27} \\
LLama 4 Scout & 0.416 $\pm$ 0.29 & 0.279 $\pm$ 0.33 & 0.391 $\pm$ 0.30 & 0.217 $\pm$ 0.30 & 0.721 $\pm$ 0.20 & 0.388 $\pm$ 0.30 & 0.715 $\pm$ 0.29 \\
Gemini 2.0 Flash & \textbf{\textcolor{gold}{0.480 $\pm$ 0.32}} & \textbf{\textcolor{gold}{0.367 $\pm$ 0.35}} & \textbf{\textcolor{gold}{0.470 $\pm$ 0.33}} & \textbf{\textcolor{gold}{0.312 $\pm$ 0.34}} & \textbf{\textcolor{gold}{0.751 $\pm$ 0.20}} & \textbf{\textcolor{gold}{0.479 $\pm$ 0.35}} & 0.719 $\pm$ 0.31 \\
Gemma 3n & \textcolor{bronze}{0.430 $\pm$ 0.31} & \textcolor{silver}{0.320 $\pm$ 0.33} & \textcolor{bronze}{0.411 $\pm$ 0.32} & \textcolor{silver}{0.250 $\pm$ 0.29} & 0.703 $\pm$ 0.23 & \textcolor{bronze}{0.419 $\pm$ 0.32} & 0.668 $\pm$ 0.33 \\
Phy 4 & \textcolor{silver}{0.435 $\pm$ 0.29} & \textcolor{bronze}{0.307 $\pm$ 0.31} & \textcolor{silver}{0.416 $\pm$ 0.30} & \textcolor{bronze}{0.247 $\pm$ 0.29} & \textcolor{silver}{0.743 $\pm$ 0.18} & \textcolor{silver}{0.451 $\pm$ 0.29} & \textbf{\textcolor{gold}{0.768 $\pm$ 0.26}} \\
Qwen3-235b & 0.368 $\pm$ 0.23 & 0.223 $\pm$ 0.24 & 0.343 $\pm$ 0.24 & 0.163 $\pm$ 0.21 & 0.720 $\pm$ 0.15 & 0.376 $\pm$ 0.25 & \textcolor{silver}{0.766 $\pm$ 0.28} \\
\bottomrule
\end{tabular}
}
\end{table}

\vspace{-0.2cm}


A métrica SBERT, que avalia a similaridade semântica entre sentenças geradas e de referência, apresentou resultados relevantes. O Gemini 2.0 Flash obteve a maior pontuação ($0.751 \pm 0.20$), seguido pelo Phi 4 ($0.748 \pm 0.18$), indicando respostas semanticamente próximas às de referência, mesmo com variações de vocabulário. As métricas ROUGE, que medem sobreposição de palavras e sequências, também foram aplicadas, com o Gemini 2.0 Flash alcançando os melhores resultados em ROUGE-L ($0.470 \pm 0.33$) e nas demais métricas (BLEU e METEOR).

Em relação à eficiência, a Tabela~\ref{tab:tempo} apresenta o tempo de resposta de cada modelo. O GPT-4o foi o mais rápido, com média de $2,101$ segundos. Os modelos LLama 4 Scout e Gemini 2.0 Flash também tiveram bom desempenho, com $2,455$ e $2,525$ segundos, respectivamente. Já o DeepSeek R1 ($11,439$~s) e o Qwen3-235b ($6,192$~s) exibiram tempos superiores, sendo que este último apresentou ainda o maior desvio padrão ($6,741$), revelando maior variabilidade no tempo de processamento.


\vspace{-0.3cm}

\begin{table}[htb]
\centering
\caption{Comparação de desempenho entre LLMs em relação ao tempo de resposta à pergunta. O melhor resultado está em \textbf{negrito}.}
\label{tab:tempo}
\footnotesize
\begin{tabular}{lcc}
\hline
Modelo                    & Média  & Desvio Padrão \\ \hline
\textbf{GPT 4o}           & \textbf{2,101 } & \textbf{0,695}        \\
\textbf{DeepSeek R1}      & 11,439 & 3,269         \\
\textbf{LLama 4 Scout}    & 2,455  & 0,784         \\
\textbf{Gemini 2.0 Flash} & 2,525  & 0,808         \\
\textbf{Gemma 3n}         & 3,734  & 2,320         \\
\textbf{Phi 4}            & 3,560  & 1,841         \\
\textbf{Qwen3-235b}       & 6,192  & 6,741         \\ \hline
\end{tabular}
\end{table}

\vspace{-0.2cm}

A análise conjunta dos resultados de qualidade e tempo de resposta revela um equilíbrio entre os modelos. Enquanto Phi 4 e Qwen3-235b se destacaram pela alta qualidade das respostas na avaliação pelo \textit{LLM-as-a-Judge}, modelos como Gemini 2.0 Flash e GPT 4o ofereceram um balanço favorável entre respostas de qualidade e maior velocidade de processamento. Ressalta-se ainda o desempenho do modelo Gemma 3n, que apresentou resultados consistentes quando às métricas e tempo razoável quando comparado a outros modelos. 

Pensando na criação de ferramentas para instituições de ensino públicas, a adoção de modelos \textit{OpenSource} se torna uma alternativa interessante, e neste sentido, os modelos Phi 4 e Gemma 3n são os mais indicados, pois permitem a execução em servidores dedicados ou localmente, sem a necessidade de pagamento sob demanda pelos serviços de geração, como no caso de plataformas como o GPT e Gemini.

\section{Considerações finais e Trabalhos futuros}\label{sec:figs}

Este trabalho abordou o desafio recorrente enfrentado por estudantes universitários para acessar informações acadêmicas, que frequentemente se encontram dispersas em múltiplos documentos e plataformas institucionais. A fragmentação da informação gera perda de tempo e insegurança quanto aos regulamentos das universidades. Para solucionar esta questão, foi desenvolvido o Sabiá, um assistente virtual que utiliza LLMs e RAG para centralizar e simplificar o acesso a esses conteúdos. Além de responder a essa necessidade, a proposta apresenta uma solução facilmente adaptável e replicável em diferentes contextos institucionais, incluindo outras universidades, escolas e secretarias de educação, contribuindo para a digitalização e modernização do ensino público brasileiro.

O projeto implementou uma arquitetura que permite a integração de diversos LLMs, o quais foram avaliados por meio de métricas de qualidade e por um LLM atuando como juiz (\textit{LLM-as-a-Judge}). Os resultados da avaliação indicaram que dentre os modelos testados, os modelos Phi 4, Gemini 2.0 Flash e Gemma 3n apresentaram resultados mais consistentes, tanto no quesito qualidade quando no tempo de resposta. Os modelos Gemma e Phi são \textit{OpenSource}, o que permite a execução destes em servidores dedicados, garantindo segurança das informações e mais controle de custos pela instituição de ensino. A análise demonstrou que é viável utilizar a GenIA junto de um \textit{pipeline} RAG para oferecer respostas rápidas e assertivas, treinada com base em documentos institucionais.

Como próximos passos, pretende-se aumentar o escopo de testes, incluindo uma maior gama de perguntas ao FAQ. Também está prevista a realização de testes empíricos com estudantes e docentes da instituição, mediante aprovação prévia em comitê de ética, para avaliar a usabilidade, a acessibilidade e a experiência de interação com o Sabiá. Além disso, é planejado a expansão da aplicação da técnica \textit{LLM-as-a-Judge}, utilizando múltiplos juízes e mais execuções para aumentar a confiabilidade dos resultados.

\bibliographystyle{sbc}
\bibliography{sbc-template}

\end{document}